\documentclass{article}

\usepackage[nonatbib]{neurips_2020}
\usepackage[utf8]{inputenc} 
\usepackage[T1]{fontenc}    
\usepackage{hyperref}       
\usepackage{url}            
\usepackage{booktabs}       
\usepackage{amsfonts}       
\usepackage{nicefrac}       
\usepackage{microtype}      
\usepackage{graphicx}
\usepackage{caption}

\title{Quantitative Assessment of Drought Impacts Using XGBoost based on the Drought Impact Reporter}

\begin{document}

\maketitle

\begin{abstract}
Under climate change, the increasing frequency, intensity, and spatial extent of drought events lead to higher socio-economic costs. However, the relationships between the hydro-meteorological indicators and drought impacts are not identified well yet because of the complexity and data scarcity. In this paper, we proposed a framework based on the extreme gradient model (XGBoost) for Texas to predict multi-category drought impacts and connected a typical drought indicator, Standardized Precipitation Index (SPI), to the text-based impacts from the Drought Impact Reporter (DIR). The preliminary results of this study showed an outstanding performance of the well-trained models to assess drought impacts on agriculture, fire, society \& public health, plants \& wildlife, as well as relief, response \& restrictions in Texas. It also provided a possibility to appraise drought impacts using hydro-meteorological indicators with the proposed framework in the United States, which could help drought risk management by giving additional information and improving the updating frequency of drought impacts. Our interpretation results using the Shapley additive explanation (SHAP) interpretability technique revealed that the rules guiding the predictions of XGBoost comply with domain expertise knowledge around the role that SPI indicators play around drought impacts.
\end{abstract}

\section{Introduction}

Drought is one of the most costly natural disasters in the world because of its broad impacts on various sectors in society \cite{hayes2012drought}. Ongoing climate change is inclined to increase the frequency and intensity of drought by raising extreme variabilities over space and time in the hydrological cycle \cite{quiggin2010drought,ding2011measuring,mukherjee2018climate}. However, compared to other natural disasters, such as floods and wildfire, drought impacts often lack structural and visible existence. Thus, drought impacts on different socio-economic aspects could be either tangible or intangible, direct or indirect \cite{ding2011measuring,van2016drought}. Additionally, based on the propagation of a drought event, its impacts could last for weeks to even years \cite{hayes2012drought}. The complex drought characteristics make it difficult to quantitatively monitor and evaluate drought impacts under climate change.

Many drought indicators have been developed to monitor drought intensity and frequency in recent decades. They are commonly grouped into meteorological, agriculture, hydrological, and composite drought indices based on the drought types \cite{mishra2010review,zargar2011review,dai2011drought}. However, only a few studies have tried to connect and calibrate drought indicators to various drought impacts, although we need to transform the temporal and spatial information of drought intensity and frequency into its impacts to help people and government agencies proactively prepare and mitigate drought \cite{bachmair2016quantitative}.

Overall, two main challenges for quantitatively evaluating drought impacts are: 1. the complexity and non-linearity of relationships between drought indicators and drought impacts, and 2. scarcity of quantitative impacts data in multiple sectors with high quality and spatial and temporal resolution. To address those challenges, several studies employed text-based data sets, such as the European Drought Impact Report Inventory (EDII), as well as various regression models, in order to estimate the drought impacts \cite{stagge2015modeling, bachmair2015exploring, blauhut2015towards,bachmair2017developing,de2020near,tadesse2015assessing}. However, only one of the studies employed a machine-learning model (Random Forest) and results indicated that it had a better performance than regression models for describing complex drought impacts \cite{bachmair2017developing}. Additionally, because of the complexity and non-linearity of drought, the observations of its impacts are associated with the social vulnerability and resilience of the local ecosystem and environment \cite{wilhite2007understanding}. Hence, an imbalanced sample distribution is common when the drought impacts data are collected, especially in the extreme categories, such as fire. Because of the potentially extreme cost of drought impacts, using a modeled approach to predict impacts will improve proactive drought response management. Overall, there is a need to develop a study with a systematic machine learning framework to link the imbalanced and multi-dimensional drought impacts with the drought indicators.

In this paper, we propose a machine-learning framework to predict multi-category drought impacts based on drought indices and impact reports in the United States, which to the best of our knowledge presents the first such attempt in the drought studies. The framework was developed based on the extreme gradient model (XGBoost) and tested by a case study in Texas during an identified drought period. We further interpret our machine learning algorithm using the Shapley additive explanation (SHAP), in order to render results that can be deemed trustworthy by domain experts.

\section{Data and Methods}

\subsection{Data}

We acquired the text-based drought impacts data set from the Drought Impact Reporter (DIR), developed and maintained by the National Drought Mitigation Center (NDMC). The DIR collects drought impacts from multiple resources and organizes them into nine categories\footnote[1]{The nine categories are agriculture, energy, plants \& wildlife, society \& public health, water supply \& quality, business \& industry, fire, relief, response \& restrictions, and tourism \& recreation}. Monthly Standardized Precipitation Index (SPI) in the Pearson Type III distribution at the various temporal scales (1, 3, 6, 9, and 12 months) were generated based on a 30-year precipitation record. The precipitation data set was acquired from the Climate Hazards Group InfraRed Precipitation with Station (CHIRPS) \cite{funk2015climate}. Seasons and months were added to the predictors for counting the seasonality and temporal trend. Additionally, to describe the spatial characteristics of drought impacts, we employed the following geographic data sets in the case study: Land Cover (LC), Public Health Regions (PHR), Regional Water Project and Development (RWPD), and Texas A\&M AgriLife Extension Service Districts (TAESD). All seasons, months, and spatial districts are categorical data sets.

\subsection{The Proposed Framework}

The framework of evaluating multi-dimensional drought impacts was developed based on the XGBoost model, combining the drought monitoring with a typical machine-learning pipeline.

\textbf{Data preparation and feature engineering.} The drought impacts from the DIR were quantitatively summarized by month and converted to dummy variables (presence versus absence). The precipitation records were aggregated monthly and calculated to SPI1-12. We applied one-hot encoding on categorical data to remove the numerical categories and their effects on the model. All data were processed at the county level and divided into training, validation, and test data sets by stratifying based on the sample distribution of the impacts. 

\textbf{Addressing imbalanced data.} For the drought impacts with a significant skew in the distribution that the proportion of the positive class is smaller than 20\%, we applied the Synthetic Minority Oversampling Technique (SMOTE) and Random Undersampling on the training data sets in order to balance the class distribution and increase the model's response to the minority class in an attempt to improve learning. This step significantly improved the F2 score and recall for the models with an imbalanced sample distribution, such as fire. We also incorporated elements of cost-sensitive learning in the training of XGBoost.

\textbf{Train and validate XGBoost models.} XGBoost is an efficient implementation of the gradient boosting decision tree that employs the second-order Taylor series to proximate the cost function and adds the regularization term in the objective function \cite{chen2016xgboost}. The reasons why we selected XGBoost to build models are five-fold. 1. Rule-based models such as decision-tree-based models are generally better suited than deep learning algorithms considering that our data set is of moderate size. 2. XGBoost models are convenient to build, in that they can attain highly-optimized performance by following standard hyperparameter search techniques implemented using stratified k-fold cross validation resampling during the training phase. XGBoost also can be easily trained in such a way to reduce overfitting. 3. XGBoost has been used successfully for winning several machine learning competitions. 4. Previous drought studies have used XGBoost for predicting meteorological indicators \cite{han2019coupling,zhang2019meteorological}. 5. XGBoost can incorporate elements of cost-sensitive learning where a cost-matrix can help influence the model to produce less false negatives. We built and trained an XGBoost model for every selected category from the text-based drought impacts data. The binary cross-entropy loss function was applied in all of the models. Additionally, we tuned the following hyperparameters: gamma and the maximum depth of the XGBoost trees to activate pruning; lambda, the L2 regularization parameter; as well as the scale of positive weight that provides cost-sensitive training. A stratified 10-fold cross validation was used to validate the model's stability on the validation data set after fine-tuning. The cross validation was designed in such a way so as to choose the hyperparameters that optimized the F2 score and area under the precision-recall curve (PR AUC). The latter two metrics emphasize that the recall values of our model are more important than precision.





\textbf{Test and interpret models.} To evaluate the models' performance, we calculated the F2 score, recall, and accuracy on the test data set. The F2 score is the F-beta measure when beta is equal to 2 so as to increase the weight of recall. The F2 score and recall were selected because we considered the false negatives more costly than the false positives in predicting drought impacts. Additionally, the SHAP was employed to estimate each feature's contribution to the model and explain the interactions among features.

Our work was written using Python 3.7 with the packages: scikit-learn, xgboost, numpy, pandas, gdal, netcdf4, xarray, geopandas, and climate\_indices.

\section{Results and Discussion}

To examine the framework, we developed a case study for one of the most severe droughts in Texas from October 2010 to June 2015\footnote[1]{The period was identified based on the time-series plot from the United State Drought Monitor.}. Energy, business and industry, and tourism and recreation were dropped from the category because they accounted for less than 5\% of the drought impacts during the period. Table 1 summarizes the proposed framework's performance for assessing multi-dimensional drought impacts on the test data set. Except for water supply and quality (0.78) and plants and wildlife (0.79), the rest models' accuracies ranged from 0.85 to 0.90. The water supply and quality model also had the lowest recall (0.51) and the F2 score (0.55). If we exclude this model, the rest of the models had recall values ranging from 0.72 to 0.96 and F2 score from 0.68 to 0.94. The fire model has the largest difference between the recall and the F2 score because we sacrificed the precision to gain a higher recall. Further studies are required to investigate why the water supply and quality model poorly performed. Overall, the framework has a good performance on evaluating various drought impacts using hydro-meteorological drought indices.

\begin{table}
  \caption{Summary of models performance on predicting drought impacts in the test data set, and the ratio of impacts is the number of impacts versus the total samples.}
  \label{sample-table}
  \centering
  \begin{tabular}{lllll}
    \toprule
    &&\multicolumn{3}{c}{Evaluation}\\
    \cmidrule(r){3-5}
    Category of Drought Impacts & Ratio of Impacts & Accuracy & Recall & F2 Score\\
    \midrule
    Agriculture & 0.69 & 0.86  & 0.93 & 0.92    \\
    Plants \& Wildlife & 0.29 & 0.79 & 0.79 & 0.74    \\
    Society \& Public Health & 0.50 & 0.90 & 0.96 & 0.94 \\
    Water Supply \& Quality & 0.36  & 0.78  & 0.51 & 0.55 \\
    Fire & 0.11  &  0.88 & 0.80 & 0.68 \\
    Relief, Response \& Restrictions & 0.36 & 0.85 & 0.72 & 0.74\\
    \bottomrule
  \end{tabular}
\end{table}

We now try to interpret the best-performing model using SHAP. Figure 1 shows the SHAP summary plot for drought indicators. Since the SHAP explainer has no explicit support at interpreting the one-hot encoded categorical data sets, we dropped them from the plot. The order of the features reveals their contributions to the model of society and public health impacts. The SPI with a 12-month moving window has the most significant impact on the model, followed by SPI6 and SPI9. SPI1 has the lowest impact on the model. Besides, most positive contributions result from the negative SPI values, except for some cases in SPI6 and SPI9, where some positive SPI values positively impact the model. This is in line with domain knowledge and expertise as follows. Typically, prolonged drought events are likely to lead to notable impacts on society and public health. Quick and minor drought events may not have any significant effects in this particular category because of the resilience in the human dimension. While SPI12 could enhance the signal of severe and prolonged droughts, one would expect it would have a more significant role in the model. However, future studies are required to explore why some higher positive SPI values would positively influence the drought impacts on society and public health. A possible inference is the quality of the drought impact data sets. Figure 2 is the SHAP main effect plot for SPI6 and SPI12 with the largest impacts on the model. The two scatter plots have a similar trend: the lowest values have the largest impacts on the model, while the impact on the model decreases with the SPI values increasing. It indicates that lower negative values in SPI6 and SPI12 would increase the probability of drought impacts on society and public health. However, further studies need to be done to explain the minor peak where SPI6 is around 0.8.



\begin{figure}
\centering
\begin{minipage}{.42\textwidth}
  \centering
  \includegraphics[scale = 0.25]{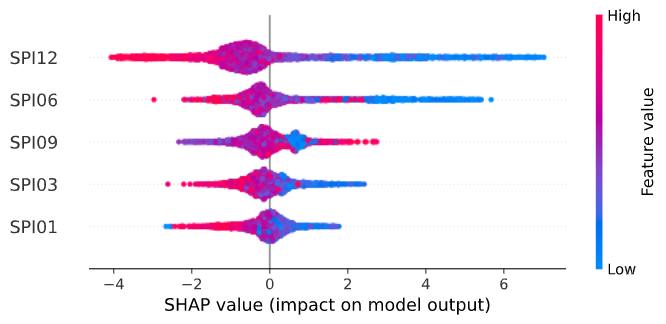}
\end{minipage}%
\begin{minipage}{.58\textwidth}
  \centering
  \includegraphics[scale = 0.32]{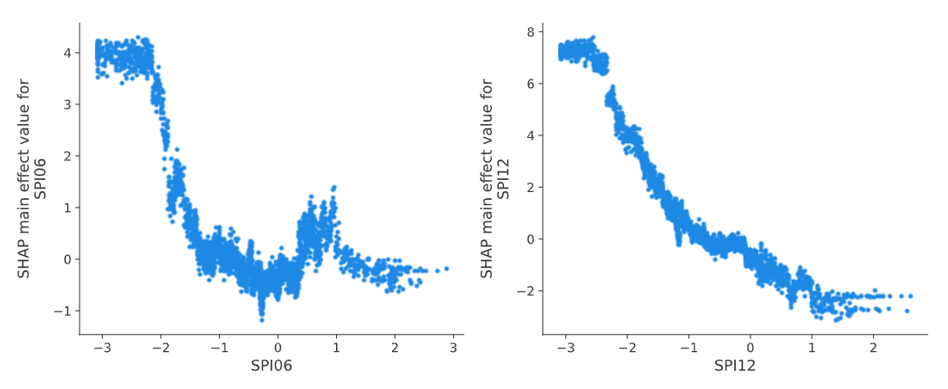}
\end{minipage}
\par
\medskip
\noindent
\begin{minipage}[t]{.42\textwidth}
  \centering
  \captionof{figure}{SHAP summary plot for SPI.}
  \label{fig:test1}
\end{minipage}%
\hfill
\begin{minipage}[t]{.58\textwidth}
  \centering
  \captionof{figure}{SHAP main effect plot for SPI6 and SPI12.}
  \label{fig:test2}
\end{minipage}
\end{figure}




\section{Conclusion and Future Work}

Quantitatively identifying the various drought impacts is always a challenge to researchers. However, it is critical to transform and connect the temporal and spatial information from hydro-meteorological drought indicators to different drought impacts. This paper proposed an XGBoost-based framework to assess multi-category drought impacts with SPI and text-based data from the DIR. The framework has a good performance on the case study in Texas. The accuracy from the models running on test data sets ranged from 0.78 to 0.90, and the F2 score was from 0.55 to 0.94. We also explained and discussed the model for the drought impact on society and public health by applying the SHAP explainer, which provides a novel insight of drought impacts on society and public health. The results reveal that SPI12 had the greatest impacts on the society and public health model, and that negative SPI6 and SPI12 values might better explain the occurrence of drought impacts on society and public health more so than other indicators.

Further studies are recommended to investigate more profound reasons and linkages among the SPI indicators and drought impacts. Future work will focus on exploring the explanation of the categorical data sets and the model outputs from the other drought impacts. It is also worth applying the proposed framework to larger spatio-temporal data sets. This study opens the door to explore more machine-learning and deep-learning methods on converting information from the drought indicators about the intensity, frequency, and spatial extent to drought impacts.

\bibliographystyle{IEEEtran}
\bibliography{reference.bib}
\small
\nocite{*}

\end{document}